\RequirePackage{amsmath}
\documentclass[runningheads]{llncs}
\usepackage{hyperref}
\usepackage{makeidx}
\usepackage{graphicx}
\graphicspath{{figures/}}
\usepackage{amsmath,amssymb} 

\usepackage{tabularx,booktabs}
\usepackage{color}

\begin{document}
	\pagestyle{headings}
	\mainmatter

	\def\GCPR18SubNumber{53}

	\title{Domain Generalization with Domain-Specific Aggregation Modules}

	\author{Antonio D'Innocente \inst{1,2} \and Barbara Caputo \inst{2}}
	\institute{Sapienza University of Rome \\ Rome, Italy \\ \email{dinnocente@diag.unroma1.it} 
	\and Italian Institute of Technology \\ Milan, Italy \\ \email{\{antonio.dinnocente,barbara.caputo\}@iit.it}}

	\maketitle

	\begin{abstract}
		Visual recognition systems are meant to work in the real world. For this to happen, they must work robustly in any visual domain, and not only on the data used during training.
        Within this context, a very realistic scenario deals with \emph{domain generalization}, i.e. the ability to build visual recognition algorithms able to work robustly in several visual domains, without having access to any information about target data statistic. This paper contributes to this research thread, proposing a deep architecture that maintains separated the information about the available source domains data while at the same time leveraging over generic perceptual information. We achieve this by introducing \emph{domain-specific aggregation modules} that through an aggregation layer strategy are able to merge generic and specific information in an effective manner. Experiments on two different benchmark databases show the power of our approach, reaching the new state of the art in domain generalization.
	\end{abstract}
    
    \section{Introduction}
	\label{sec:introduction}

As artificial intelligence, fueled by machine and deep learning, is entering more and more into our everyday lives, there is a growing need for visual recognition algorithms able to leave the controlled lab settings and work robustly in the wild. This problem has long been investigated in the community under the name of Domain Adaptation (DA): considering the underlying statistics generating the data used during training (source domain), and those expected at test time (target domain), DA assumes that the robustness issues are due to a covariate shift among the source and target distributions, and it attempts to align such distributions so to increase the recognition performances on the target domain. Since its definition \cite{Saenko:2010}, the vast majority of works has focused on the scenario where one single source is available at training time, and one specific target source is taken into consideration at test time, with or without any labeled data (for an overview of previous work we refer to section \ref{sec:rel-works}). Although useful, this setup is somewhat limited: given the large abundance of visual data  produced daily worldwide and uploaded on the Web, it is very reasonable to assume that several source domains might be available at training time. Moreover, the assumption to have access to data representative of the underlying statistic of the target domain, regardless of annotation, is not always realistic. Rather than equipping a seeing machine with a DA algorithm able to solve the domain gap for a specific single target, one would hope to have methods able to solve the problem for \emph{any} target domain. This last scenario, much closer to realistic settings, goes under the name of Domain Generalization (DG, \cite{li2017deeper}), and is the focus of our work. 

Current approaches to DG tend to follow two alternative routes: the first tries to use all source data together in order to learn a joint, general representation for the categories of interest strong enough to work on any target domain \cite{MLDG_AAA18}. The second instead opts for keeping separated the information coming from each source domain, trying to estimate at test time the similarity between the target domain represented by the incoming data and the known sources, and use only the classifier branch trained on that specific source for classification \cite{mancini2018best}. Our approach sits across these two philosophies, attempting to get the best of both worlds. Starting from a generic convnet, pre-trained  on a general knowledge database like ImageNet \cite{ILSVRC15}, we build a new multi-branch architecture with as many branches as the source domains available at training time. Each branch leverages over the general knowledge contained into the pre-trained convnet through a deep layer aggregation strategy inspired by \cite{yu2017deep}, that we call Domain-Specific Aggregation Modules (D-SAM). The resulting architecture is trained so that all three branches contribute to the classification stage through an aggregation strategy. The resulting convnet can be used in an end-to-end fashion, or its learned representations can be used as features in a linear SVM. 
We tested both options on two different architectures and two different domain generalization databases, benchmarking against all recent approaches to the problem. Results show that our D-SAM architecture, in all cases, consistently achieve the state of the art.

    \section{Related Works}
	\label{sec:rel-works}

Most of work in DA has focused on single source scenarios, with two main research threads. The first deals with features, aiming to learn deep domain representations that are invariant to the domain shift, although discriminative enough to perform well on the target \cite{carlucci2017auto,carlucci2017just}, \cite{LongZ0J17}, \cite{Sun:CORAL:AAAI16}.
Other methods rely on adversarial loss functions \cite{Ganin:DANN:JMLR16}, \cite{Tzeng_ICCV2015}, \cite{sankaranarayanan2017generate}. 
Also two-step networks have been shown to 
have practical advantages \cite{Hoffman:Adda:CVPR17,LOAD_ICRA}.
The second thread 
 focuses on images.
The adversarial approach used successfully for feature-based methods, has also
been applied directly to the reduction of the visual domain gap. Various GAN-based  
strategies \cite{Goodfellow:GAN:NIPS2014} have been proposed for generating new images and/or perturb existing ones to mimic the visual style of a domain and reducing the discrepancy at the pixel level \cite{Bousmalis:Google:CVPR17}, \cite{appleGAN}, \cite{russo17sbadagan}.
Recently, some authors addressed the multi-source domain adaptation problem with deep networks.
 The approach proposed in \cite{cocktail_CVPR18} builds over \cite{Ganin:DANN:JMLR16}
by replicating the adversarial domain discriminator branch for each available source. Moreover
these discriminators are also used to get a perplexity score that indicates how the multiple
sources should be combined at test time as in \cite{Mansour_NIPS09}. 
A similar multi-way adversarial strategy is used also in \cite{MDAN_ICLRW18}, but this work comes 
with a theoretical support that frees it from the need of respecting a specific optimal source 
combination and thus from the need of learning the source weights.

In the DG setting, access to the target data is not allowed, thus the
main objective  
is to look across multiple sources for shared 
factors in the hypothesis that they will hold also for any new target domain. 
Deep DG methods are presented in \cite{mancini2018best,MassiRAL} and \cite{li2017deeper}. 
The first works propose a weighting procedure on the source models, while the second aims at separating 
the source knowledge into domain-specific and domain-agnostic sub-models.
A meta-learning approach was recently presented in \cite{MLDG_AAA18}: 
it starts by creating virtual testing domains within each source mini-batch and then it trains
a network to minimize the classification loss, while also ensuring that the taken direction leads 
to an improvement on the virtual testing loss.

Over the last years, it has emerged a growing interest on studying modules and connectivity patterns, and on how to assemble them systematically. Some studies showed how skipping connections can be beneficial for classification and regression. In particular, \cite{huang2017densely} showed how skipping connections concatenating all the layers in stages is effective for semantic fusion, while \cite{lin2017feature} exploited conceptually similar ideas for spatial fusion. An unifying framework for these approaches, on which to some extent we build, has been recently proposed in \cite{yu2017deep}. There the authors proposed two general structures for deep layer aggregation, one iterative and one hierarchical, that capture the nuances of previous works while being applicable in principle to any convnet. 

We leverage on this work, proposing a variant of iterative deep aggregation leading to a multi branch architecture, able to conjugate the need for general representations while retaining the strength of keeping information from different sources separated in the domain generalization setting.

    \section{Domain Specific Aggregation Modules}
	\label{sec:methodology}
    
In this section we describe our aggregation strategy for DG.    
We will assume to have $S$ source domains and $T$ target domains,  denoting with $N_{i}$ the cardinality of the $i_{th}$ source domain, for which we have $\{x_{j}^{i},y_{j}^{i}\}_{j=1}^{N_{i}}$ labeled samples. Source and target domains share the same classification task; however, unlike DA, the target distribution is unknown and the algorithm is expected to generalize  to new domains without ever having access to target data, and hence without any possibility to estimate the underlying statistic for the target domain.
    
The most basic approach, \emph{Deep All}, consists of ignoring the domain membership of the  images available from all training sources, and training a generic algorithm on the combined source samples. Despite its simplicity, \emph{Deep All} outperforms many engineered methods in domain generalization, as shown in \cite{li2017deeper}. 
    The domain specific aggregation modules we propose can be seen as a way to augment the generalization abilities of given CNN architectures by maintaining a generic core, while at the same time explicitly modeling the single domain specific features separately, in a whole coherent structure.
    
    Our architecture consists of a main branch $\Theta$ and a collection of domain specific aggregation modules $\Lambda = \{\lambda_{1}...\lambda_{n}\}$, each specialized on a single source domain. The main branch $\Theta$ is the backbone of our model, and it can be in principle any pre-trained off-the shelf convnet. 
 Aggregation modules, which we design inspired by an iterative aggregation protocol described in \cite{yu2017deep}, 
 receive inputs from $\Theta$ and learn to combine features at different levels to produce classification outputs. At training time, each domain-specific aggregation module learns to specialize on a single source domain. In the validation phase, we use a variation of a leave-one-domain-out strategy: we average predictions of each module but, for each ${i_{th}}$ source domain, we exclude the corresponding domain-specific module $\lambda_{i}$ from the evaluation. We test the model in both an end-to-end fashion and by running a linear classifier on the extracted features. In the rest of the section we describe into detail the various components of our approach (section \ref{sec:aggregation}-\ref{sec:architecture}) and the training protocol (section \ref{sec:learning}).

    \subsection{Aggregation Module}
    \label{sec:aggregation}
    
    \begin{figure}[t]
	\centering
		\includegraphics[width=0.8\linewidth]{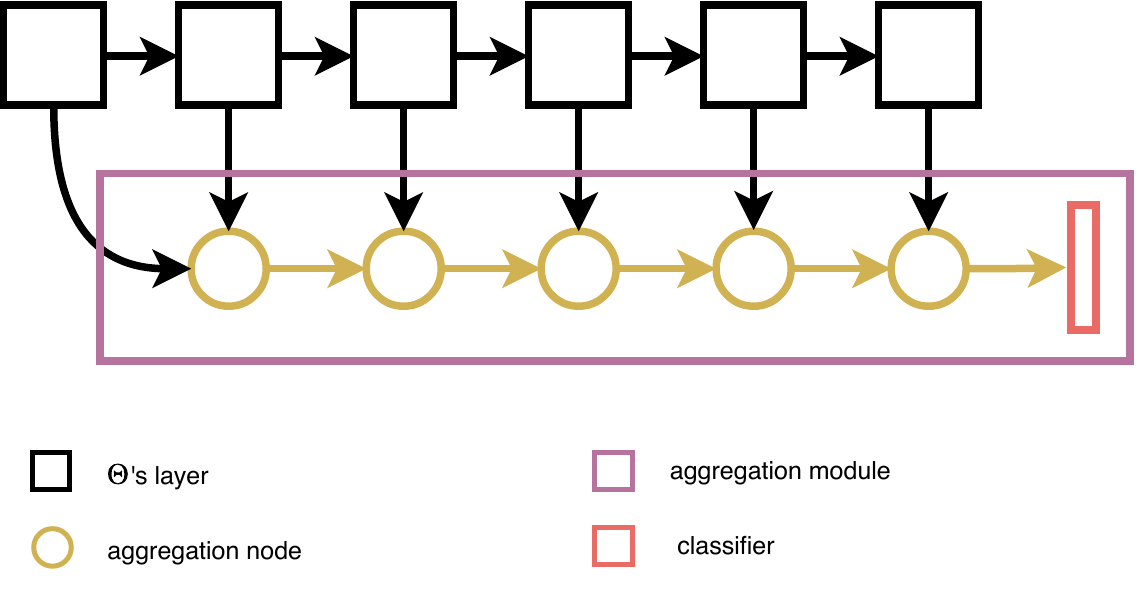}
		\caption{Architecture of an aggregation module (purple) augmenting a CNN model. Aggregation nodes (yellow) iteratively process input from $\Theta$'s layers and propagate them to the classifier.}
		\label{fig:aggr}
	\end{figure}
    
    Deep Layer Aggregation \cite{yu2017deep} is a feature fusion strategy designed to augment a fully convolutional architecture with a parallel, layered structure whose task is to better process and propagate features from the original network to the classifier. Aggregation nodes, the main building block of the augmenting structure, learn to combine convolutional outputs from multiple layers with a compression technique, which in \cite{yu2017deep} is implemented with 1x1 convolutions followed by batch normalization and nonlinearity. The arrangement of connections between aggregation nodes and the augmented network's original layers yields an architecture more capable of extracting the full spectrum of spatial and semantical informations from the original model \cite{yu2017deep}.
    
    Inspired by the aggregations of \cite{yu2017deep}, we implement aggregation modules as parallel feature processing branches pluggable in any CNN architecture. Our aggregation consists of a stacked sequence of aggregation nodes $N$, with each node iteratively combining outputs from $\Theta$ and from the previous node, as shown in Figure \ref{fig:aggr}. The nodes we use are implemented as 1x1 convolutions followed by nonlinearity. Our aggregation module visually resembles the Iterative Deep Aggregation (IDA) strategy described in \cite{yu2017deep}, but the two are different. IDA is an aggregation pattern for merging different scales, and is implemented on top of a hierarchical structure. Our aggregation module is a pluggable augmentation which merges features from various layers sequentially. Compared to \cite{yu2017deep}, our structure can be merged with any existing pre-trained model without disrupting the original features' propagation. We also extend its usage to non-fully convolutional models by viewing 2-dimensional outputs of fully connected nodes as 4-dimensional (N x C x H x W) tensors whose H and W dimension are collapsed. As we designed these modules having in mind the DG problem and their usage for domain specific learning, we call them Domain-Specific Aggregation Modules (D-SAM).

    \subsection{D-SAM Architecture for Domain Generalization}
    \label{sec:architecture}
    
    The modular nature of our D-SAMs allows the stacking of multiple augmentations on the same backbone network. Given a DG setting in which we have $S$ source domains, we choose a pre-trained model $\Theta$ and  augment it with $S$ aggregation modules, each of which implements its own classifier while learning to specialize on an individual domain. The overall architecture is shown in Figure \ref{fig:train}. 
    
    Our intention is to model the domain specific part and the domain generic part within the architecture. While aggregation modules are domain specific, we may see $\Theta$ as the domain generic part that, via backpropagation, learns to yield general features which aggregation modules specialize upon. Although not explicitly trained to do so, our feature evaluations suggest that thanks to our training procedure, the backbone $\Theta$ implicitly learns more domain generic representations compared to the corresponding backbone model trained without aggregations.
    
    \subsection{Training and Testing}
    \label{sec:learning}
    
    \begin{figure}[t]
	\centering
		\includegraphics[width=1\linewidth]{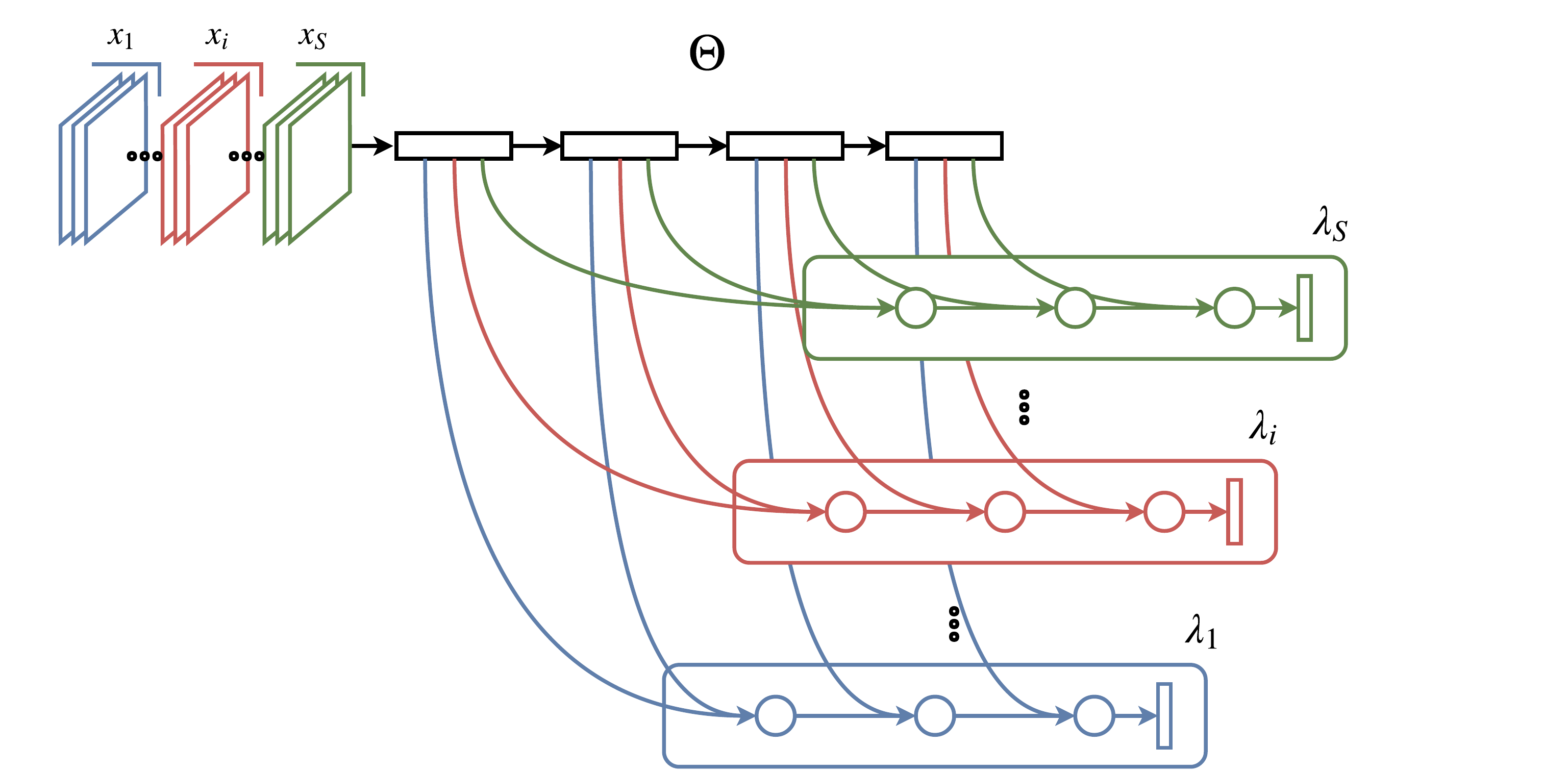}
		\caption{Simplified architecture with 3 aggregation nodes per aggregation module. The main branch $\Theta$ shares features with $S$ specialized modules. At training time, the $i_{th}$ aggregation module only processes outputs relative to the $i_{th}$ domain.}
		\label{fig:train}
	\end{figure}
    
    We train our model so that the backbone $\Theta$ processes all the input images, while each aggregation module learns to specialize on a single domain. To accomplish this, at each iteration we feed to the network $S$ equal sized mini-batches grouped by domain. Given an input mini-batch $x_{i}$ from the $i_{th}$ source domain, the corresponding output of our function, as also shown graphically in Figure \ref{fig:train}, is:
    
    \begin{equation}
    f(x_{i}) = \lambda_{i}(\Theta(x_{i})).
    \end{equation}
    
    We optimize our model by minimizing the cross entropy loss function $L_{C} = \displaystyle\sum_{c}y_{x_{j}^{i},c}\log(p_{x_{j}^{i},c})$, which for a training iteration we formalize as:
    
    \begin{equation}
    L(\Theta, \Lambda) = \displaystyle\sum_{i=1}^{S}L_{C}((\lambda_{i}\circ\Theta)(x_{i})).
    \end{equation}
    
    We validate our model by combining probabilities of the outputs of aggregation modules. One problem of the DG setting is that performance on the validation set is not very informative, since accuracy on source domains doesn't give much indication of the generalization ability. We partially mitigate this problem in our algorithm by calculating probabilities for validation as:
    
    \begin{equation}
    p_{x_{j}^v} = \sigma(\displaystyle\sum_{i=1, i\neq v}^{S}\lambda_{i}(\Theta(x_{j}^v))),
    \end{equation}
    
    where $\sigma$ is the softmax function. Given an input image belonging to the $k$ source domain, all aggregation modules besides $\lambda_{k}$ participate in the evaluation. With our validation we keep the model whose aggregation modules are general enough to distinguish between unseen distributions, while still training the main branch on all input data. 
    
    We test our model both in an end to end fashion and as a feature extractor. For end-to-end classification we calculate probabilities for the label as:
    
    \begin{equation}
    p_{x_{j}^t} = \sigma(\displaystyle\sum_{i=1}^{S}\lambda_{i}(\Theta(x_{j}^t))),
    \end{equation}
    
    When testing our algorithm as a feature extractor, we evaluate $\Theta$'s and $\Lambda$'s features by running an SVM Linear Classifier on the DG task.
    
    \section{Experiments}
\label{sec:expers}

In this section we report experiments assessing the effectiveness of our DSAM-based architecture in the DG scenario, using two different backbone architectures (a ResNet-18 \cite{he2016deep} and an AlexNet \cite{krizhevsky2012imagenet}), on two different databases. We first describe the datasets used (section \ref{sec:dataset}), and then we proceed to report the model setup (section \ref{sec:model}) and the training protocol adopted (section \ref{sec:training}). Section \ref{sec:results} reports and comments upon the experimental results obtained.

    \subsection{Datasets}
    \label{sec:dataset}

We performed experiments on two different databases. The \textbf{PACS} database \cite{li2017deeper} has been recently introduced to support research on DG, and it is quickly becoming the standard reference benchmark for this research thread. It consists of $9.991$ images, of resolution $227\times 227$, taken from four different visual domains (Photo, Art paintings, Cartoon and Sketches), depicting seven categories. We followed the experimental protocol of \cite{li2017deeper} and trained our models considering three domains as source datasets and the remaining one as target.

The \textbf{Office-Home} dataset \cite{venkateswara2017Deep} was introduced to support research on DA for object recognition. It provides images from four different domains: Artistic images, Clip art, Product images and Real-world images. Each domain depicts $65$ object categories that can be found typically in office and home settings. We are not aware of previous work using the Office-Home dataset in DG scenarios, hence we decided to follow also here the experimental setup introduced in \cite{li2017deeper} and described above for PACS.

    \begin{figure}
\centering
\begin{minipage}{.5\textwidth}
  \centering
  \includegraphics[width=.2\textwidth]{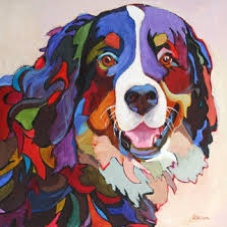}
\includegraphics[width=.2\textwidth]{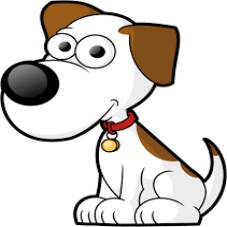}
\includegraphics[width=.2\textwidth]{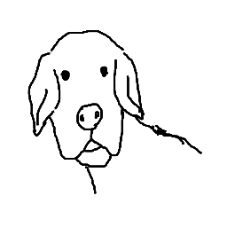}
\includegraphics[width=.2\textwidth]{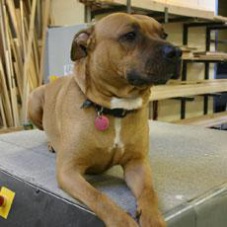}
\includegraphics[width=.2\textwidth]{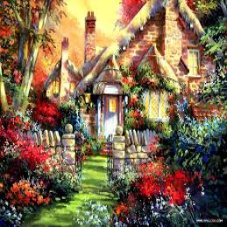}
\includegraphics[width=.2\textwidth]{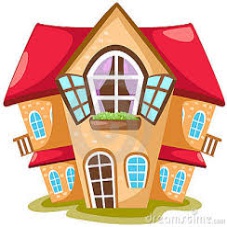}
\includegraphics[width=.2\textwidth]{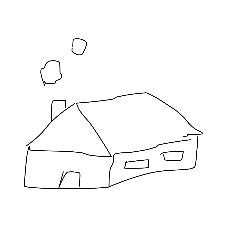}
\includegraphics[width=.2\textwidth]{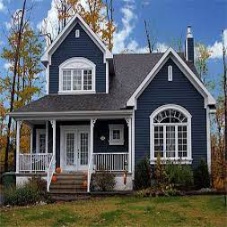}
\includegraphics[width=.2\textwidth]{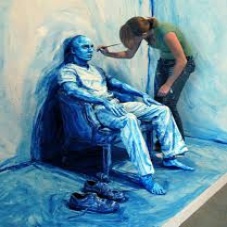}
\includegraphics[width=.2\textwidth]{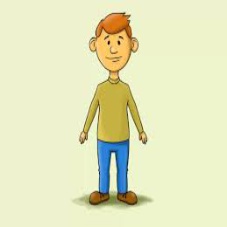}
\includegraphics[width=.2\textwidth]{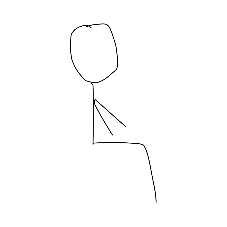}
\includegraphics[width=.2\textwidth]{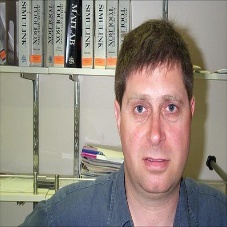}
\includegraphics[width=.2\textwidth]{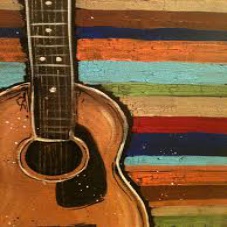}
\includegraphics[width=.2\textwidth]{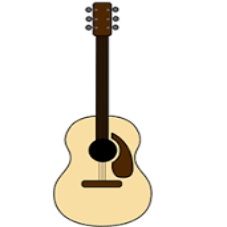}
\includegraphics[width=.2\textwidth]{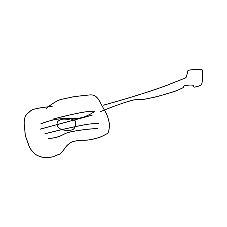}
\includegraphics[width=.2\textwidth]{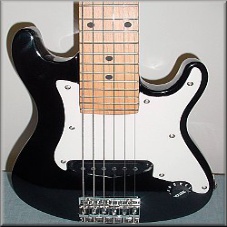}
\end{minipage}%
\begin{minipage}{.5\textwidth}
  \centering
  \includegraphics[width=.2\textwidth]{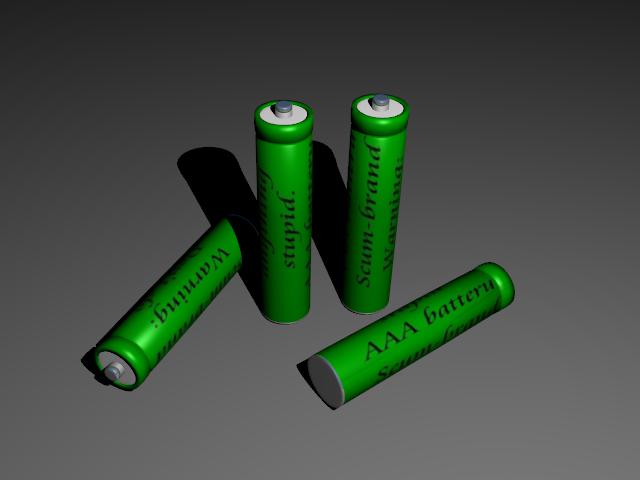}
\includegraphics[width=.2\textwidth]{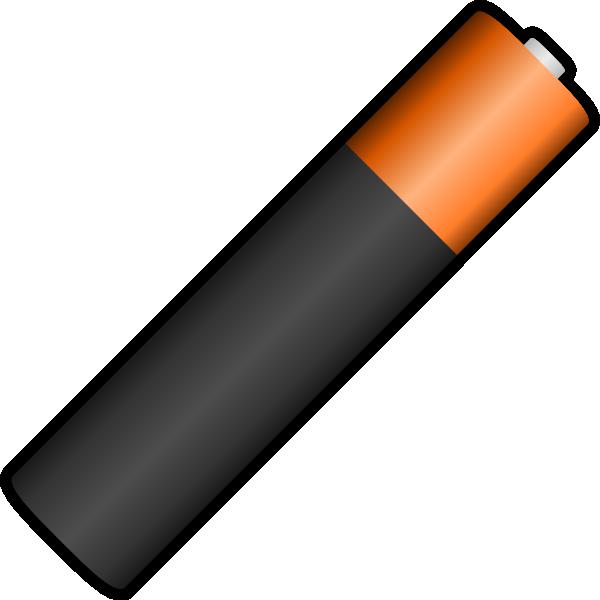}
\includegraphics[width=.2\textwidth]{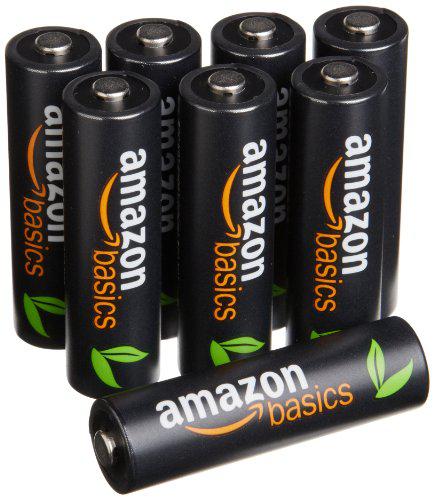}
\includegraphics[width=.2\textwidth]{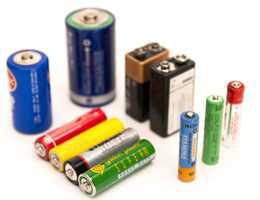}
\includegraphics[width=.2\textwidth]{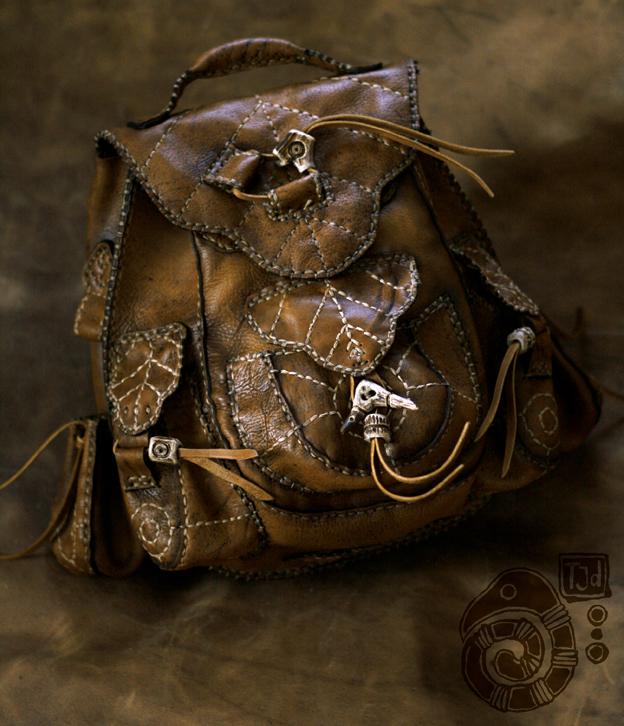}
\includegraphics[width=.2\textwidth]{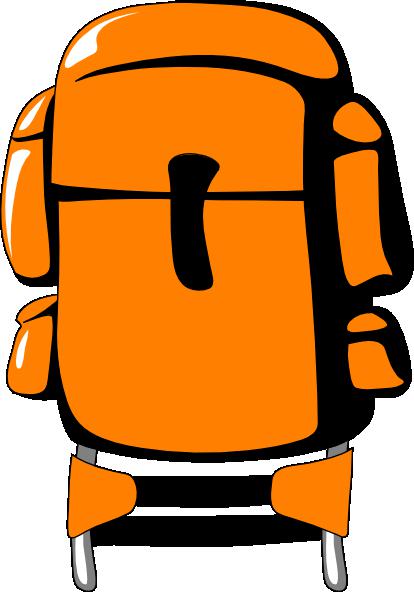}
\includegraphics[width=.2\textwidth]{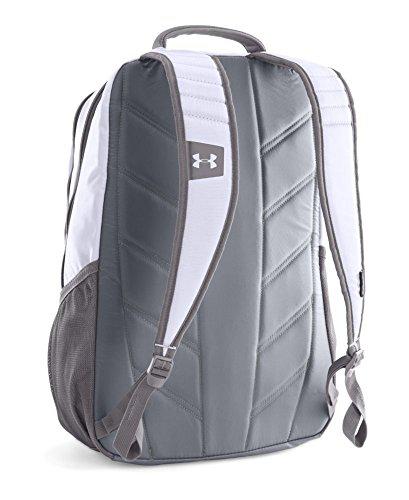}
\includegraphics[width=.2\textwidth]{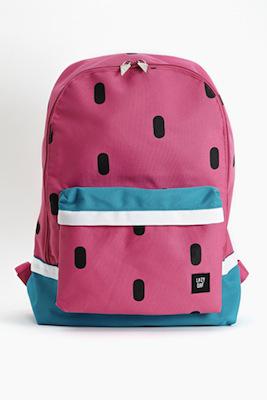}
\includegraphics[width=.2\textwidth]{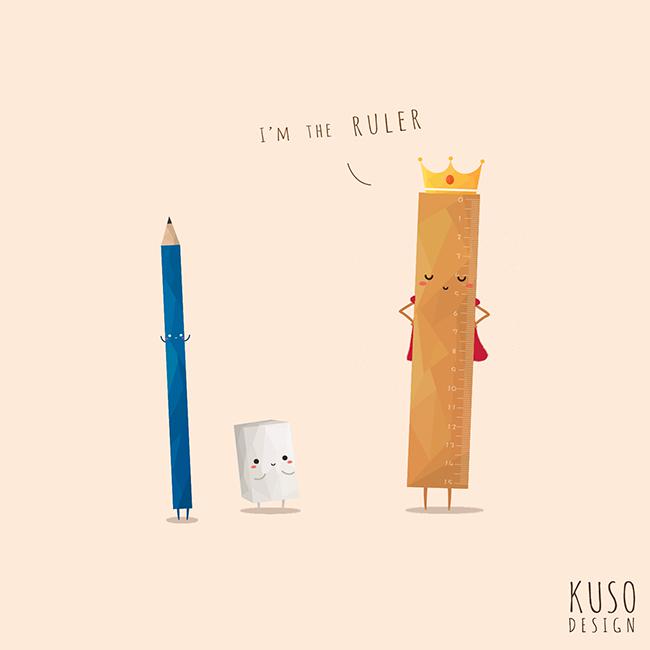}
\includegraphics[width=.2\textwidth]{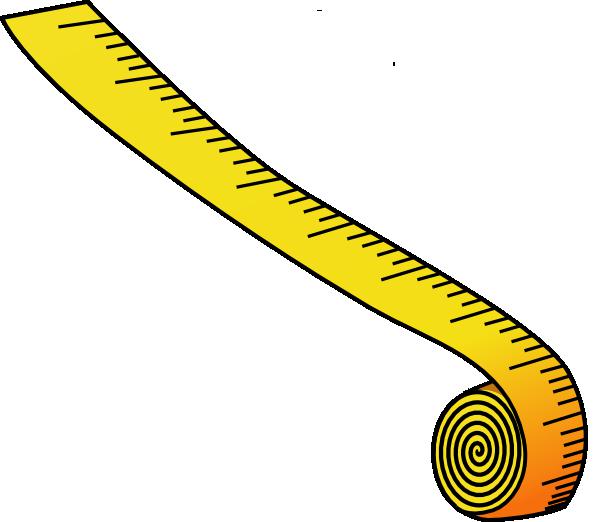}
\includegraphics[width=.2\textwidth]{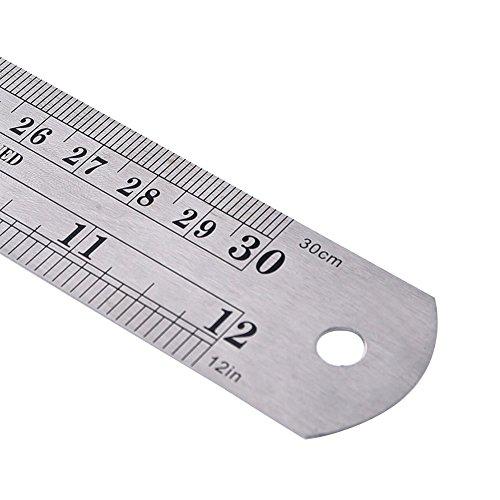}
\includegraphics[width=.2\textwidth]{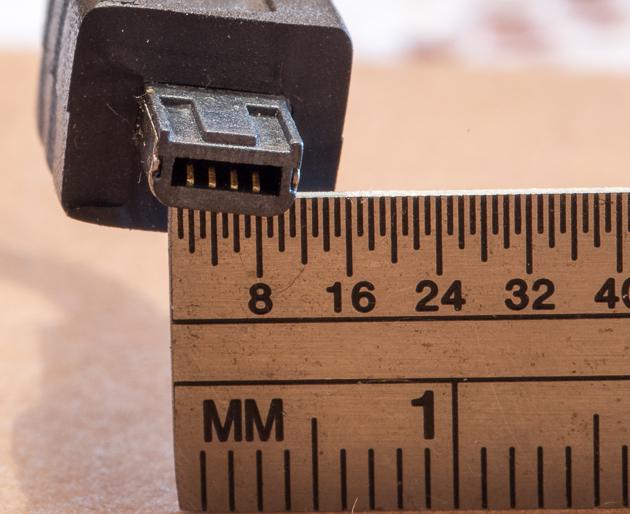}
\includegraphics[width=.2\textwidth]{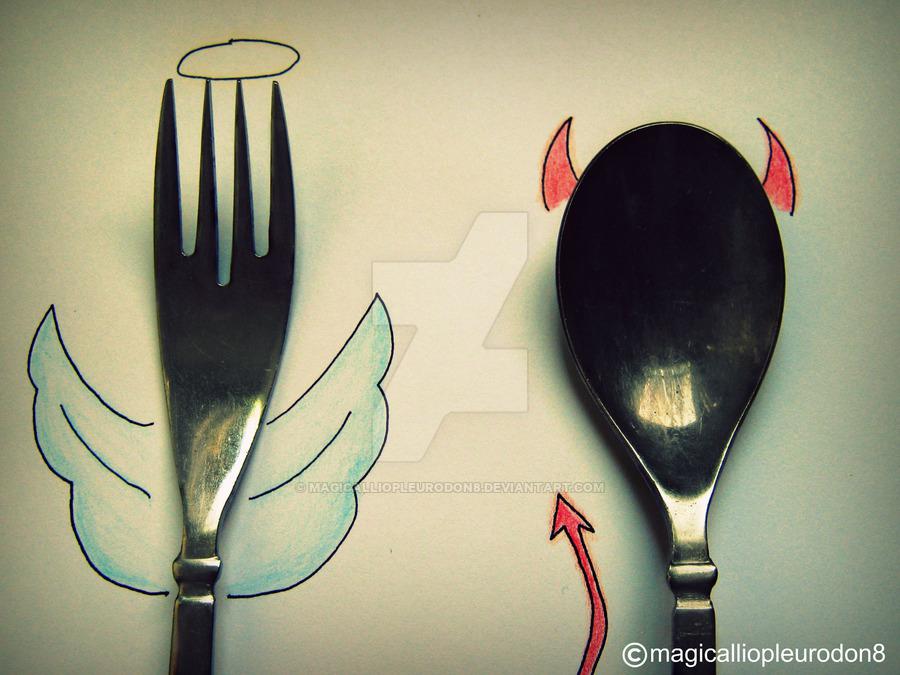}
\includegraphics[width=.2\textwidth]{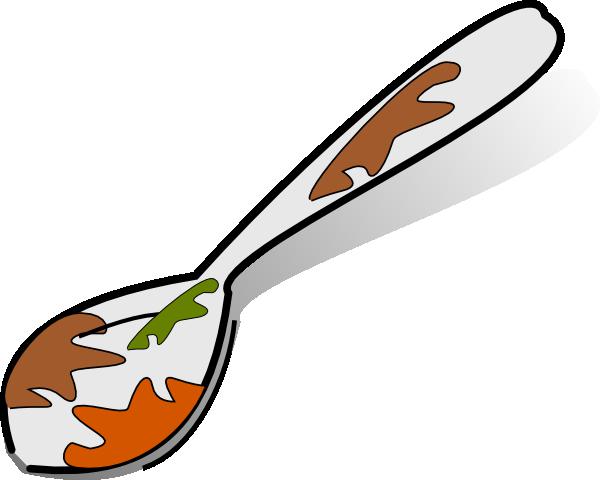}
\includegraphics[width=.2\textwidth]{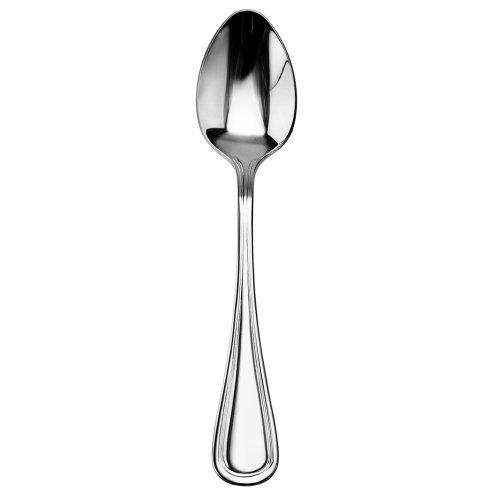}
\includegraphics[width=.2\textwidth]{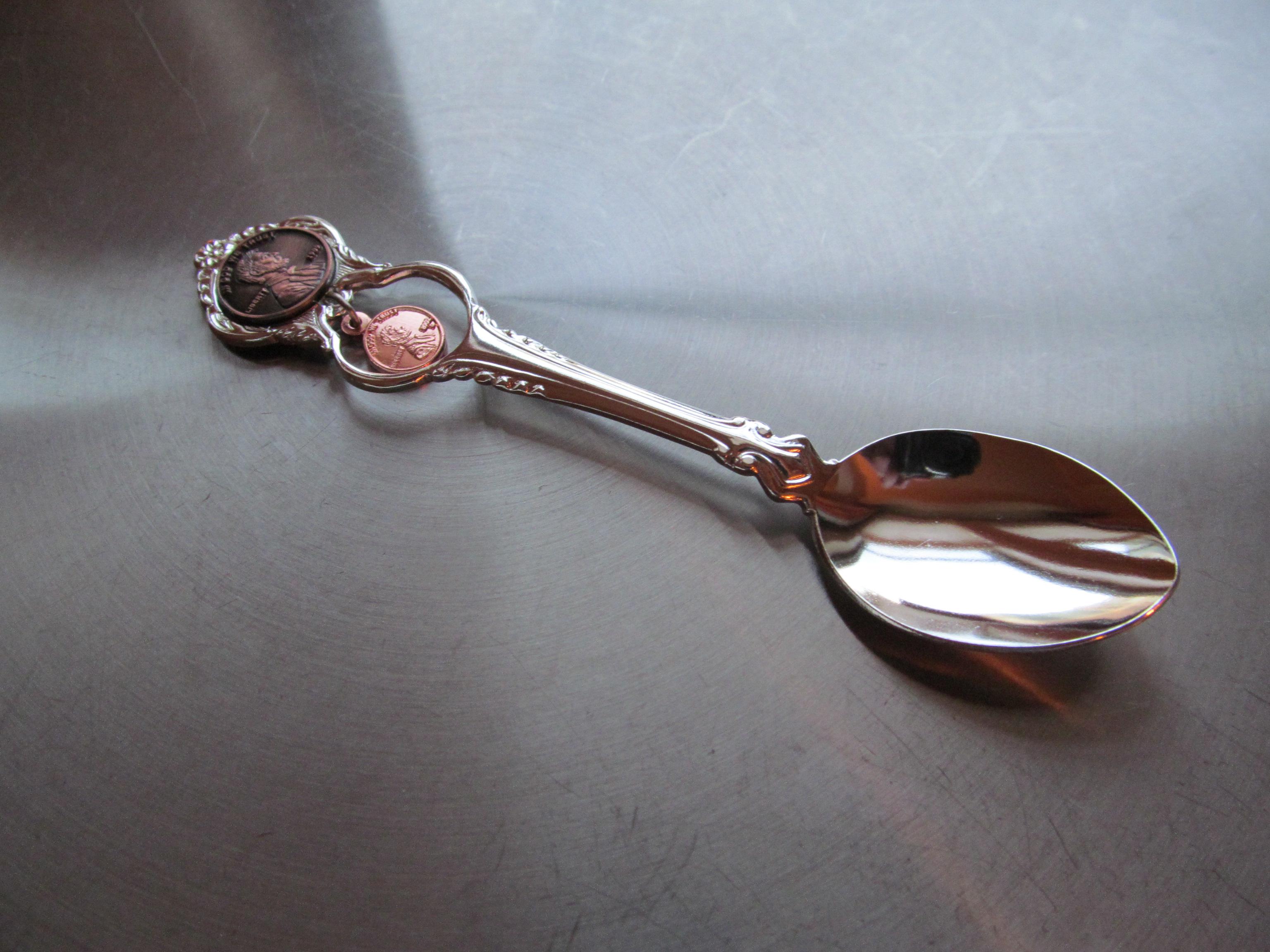}
\end{minipage}
\caption{Exemplar images for the PACS (left) and Office-Home (right) databases, on selected categories. We see that for both databases, the variations among domains for the same category can vary a lot.}
\end{figure}
    
    \subsection{Model setup}
	\label{sec:model}

	\textbf{Aggregation Nodes}. We implemented the aggregation nodes as 1x1 convolutional filters followed by nonlinearity. Compared to \cite{yu2017deep}, we did not use batch normalization in the aggregations, since we empirically found it detrimental for our difficult DG targets. Whenever the inputs of a node have different scales, we downsampled with the same strategy used in the backbone model. For ResNet-18 experiments, we further regularized the convolutional inputs of our aggregations with dropout.
    
    \textbf{Aggregation of Fully Connected Layers}. We observe that a fully connected layer's output can be seen as a 4-dimensional (N, C, H, W) tensor with collapsed height and width dimensions, as each unit's output is a function of the entire input image. A 1x1 convolutional layer whose input is such a tensor coincides with a fully connected layer whose input is a 2-dimensional (N, C) tensor, so for simplicity we implemented those aggregations with fully connected layers instead of convolutions.
    
    \textbf{Model Initialization}. We experimented with two different backbone models: AlexNet and ResNet-18, both of which are pre-trained on the ImageNet 1000 object categories \cite{ILSVRC15}. We initialized our aggregation modules  $\Lambda$ with random uniform initialization. We connected the aggregation nodes with the output of the AlexNet's layers when using AlexNet as backbone, or with the exit of each residual block when using ResNet-18. 
    
    \subsection{Training setup}
    \label{sec:training}
    We finetuned our models on $S=3$ source domains and tested on the remaining target. We splitted our training sets in 90\% train and 10\% validation, and used the best performing model on the validation set for the final test, following the validation strategy described in Section \ref{sec:methodology}. For preprocessing, we used random zooming with rescaling, horizontal flipping, brightness/contrast/saturation/hue perturbations and normalization using ImageNet's statistics. We used a batch size of 96 (32 images per source domain) and trained using SGD with momentum set at 0.9 and initial learning rate at 0.01 and 0.007 for ResNet's and AlexNet's experiments respectively. We considered an epoch as the minimum number of steps necessary to iterate over the largest source domain and we trained our models for 30 epochs, scaling the learning rate by a factor of 0.2 every 10 epochs. We used the same setup to train our ResNet-18 Deep All baselines. We repeated each experiment 5 times, averaging the results.
    
    \subsection{Results}
    \label{sec:results}
    
    We run a first set of experiments with the D-SAMs using an AlexNet as backbone, to compare our results with those reported in the literature by previous works, as AlexNet has been so far the convnet of choice in DG. Results are reported in table \ref{table:pacs}. We see that our approach outperforms previous work by a sizable margin, showing the value of our architecture. Particularly, we underline that D-SAMs obtain remarkable performances on the challenging setting where the 'Sketch' domain acts as target.
    
    We then run a second set of experiments, using both the PACS and Office-Home dataset, using as backbone architecture a ResNet-18. The goal of this set of experiments is on one side to showcase how our approach can be easily used with different $\Theta$ networks, on the other side to perform an ablation study with respect to the possibility to use D-SAMs not only in an end-to-end classification framework, but also to learn feature representations, suitable for domain generalization. To this end, we report results on both databases using the end-to-end approach tested in the AlexNet experiments, plus results obtained using the feature representations learned by $\Theta$, $\Lambda$ and the combination of the two. Specifically, we extract and $l_2$ normalize features from 
the last pooling layer of each component. We integrate features of $\Lambda$’s modules 
with concatenation, and train the SVM classifier leaving the hyperparameter $C$ 
at the default value. Our results in table \ref{table:resnetpacs} and \ref{table:resnetoh} show that the SVM classifier 
trained on the $l_2$ normalized features always outperforms the corresponding end-
 to-end models, and that $\Theta$’s and $\Lambda$’s features have similar performance, with $\Theta$’s 
 features outperforming the corresponding Deep All features while requiring no 
 computational overhead for inference.

    \begin{table}
\centering
\caption{PACS end-to-end results using D-SAMs coupled with the AlexNet architecture.}
\label{table:pacs}
\begin{tabular}{p{2cm}p{2cm}p{2cm}p{2cm}p{2cm}p{2cm}}
\hline
\textbf{}    & Deep All \cite{li2017deeper} & TF \cite{li2017deeper} & MLDG \cite{MLDG_AAA18}           & SSN \cite{mancini2018best}         & D-SAMs           \\
\hline
art painting & 64.91    & 62.86        & \textbf{66.23} & 64.10          & 63.87          \\
cartoon      & 64.28    & 66.97        & 66.88          & 66.80          & \textbf{70.70} \\
photo        & 86.67    & 89.50        & 88.00          & \textbf{90.20} & 85.55          \\
sketch       & 53.08    & 57.51        & 58.96          & 60.10          & \textbf{64.66} \\
\hline
avg          & 67.24    & 69.21        & 70.01          & 70.30          & \textbf{71.20}
\end{tabular}
\end{table}

\begin{table}
\centering
\caption{PACS results with ResNet-18 using features (top-rows) and end-to-end accuracy (bottom rows).}
\label{table:resnetpacs}
\begin{tabular}{p{2.5cm}p{1.9cm}p{1.9cm}p{1.9cm}p{1.9cm}|p{1.9cm}}
\hline
\textbf{}   & art painting & cartoon & sketch & photo & Avg   \\
\hline
Deep All (feat.) & 77.06        & \textbf{77.81}   & 74.09  & 93.28 & 80.56 \\
$\Theta$ (feat.)    & \textbf{79.57}        & 76.94   & 75.47  & 94.16 & 81.54 \\
$\Lambda$ (feat.)    & 79.48        & 77.13   & 75.30  & 94.30 & \textbf{81.55} \\
$\Theta + \Lambda$ (feat.) & 79.44 & 77.22 & 75.33 & 94.19 & 81.54
\\
\hline
Deep All    & 77.84        & 75.89   & 69.27  & 95.19  & 79.55 \\
D-SAMs       & 77.33        & 72.43   & \textbf{77.83}  &  \textbf{95.30} & 80.72
\end{tabular}
\end{table}

\begin{table}
\caption{OfficeHome results with ResNet-18 using features (top rows) and end-to-end accuracy (bottom rows).}
\label{table:resnetoh}
\begin{tabular}
{p{2.5cm}p{1.9cm}p{1.9cm}p{1.9cm}p{1.9cm}|p{1.9cm}}
\hline
\textbf{}      & Art   & Clipart & Product & Real-World & Avg   \\
\hline
Deep All (feat.) & 52.66 & 48.35   & 71.37   & 71.47      & 60.96 \\
$\Theta$ (feat.) & 54.55 & \textbf{49.37}   & 71.38   & \textbf{72.17}      & \textbf{61.87} \\
$\Lambda$ (feat.)     & 54.53    & 49.04      & 71.57      & 71.90         & 61.76    \\
$\Theta + \Lambda$ (feat.) & 54.54 & 49.05 & \textbf{71.58} & 72.03 & 61.80 \\
\hline
Deep All       & 55.59 & 42.42   & 70.34   & 70.86      & 59.81 \\
D-SAMs          & \textbf{58.03} & 44.37   & 69.22   & 71.45      & 60.77
\end{tabular}
\end{table}
    \section{Conclusions}
This paper presented a Domain Generalization architecture inspired by recent work on deep layer aggregation. We developed a convnet that, starting from a pre-trained model carrying generic perceptual knowledge, aggregates layers iteratively for as many branches as the available source domains data at training time. The model can be used in an end-to-end fashion, or its convolutional layers can be used as features in a linear SVM. Both approaches, tested with two popular pre-trained architectures on two benchmark databases, achieve the new state of the art.
Future work will further study deep layer aggregation strategies within the context of domain generalization, as well as scalability with respect to the number of sources.

	\bibliographystyle{splncs04}
	\bibliography{egbib}

\begin{thebibliography}{10}
\providecommand{\url}[1]{\texttt{#1}}
\providecommand{\urlprefix}{URL }
\providecommand{\doi}[1]{https://doi.org/#1}

\bibitem{LOAD_ICRA}
Angeletti, G., Caputo, B., Tommasi, T.: Adaptive deep learning through visual
  domain localization. In: International Conference on Robotic Automation
  (ICRA) (2018)

\bibitem{Bousmalis:Google:CVPR17}
Bousmalis, K., Silberman, N., Dohan, D., Erhan, D., Krishnan, D.: Unsupervised
  pixel-level domain adaptation with gans. In: Computer Vision and Pattern
  Recognition (CVPR) (2017)

\bibitem{carlucci2017auto}
Carlucci, F.M., Porzi, L., Caputo, B., Ricci, E., Rota~Bul{\`o}, S.: Autodial:
  Automatic domain alignment layers. In: International Conference on Computer
  Vision (ICCV) (2017)

\bibitem{carlucci2017just}
Carlucci, F.M., Porzi, L., Caputo, B., Ricci, E., Rota~Bul{\`o}, S.: Just dial:
  domain alignment layers for unsupervised domain adaptation. In: International
  Conference on Image Analysis and Processing (2017)

\bibitem{Ganin:DANN:JMLR16}
Ganin, Y., Ustinova, E., Ajakan, H., Germain, P., Larochelle, H., Laviolette,
  F., Marchand, M., Lempitsky, V.: Domain-adversarial training of neural
  networks. J. Mach. Learn. Res.  \textbf{17}(1),  2096--2030 (2016)

\bibitem{Goodfellow:GAN:NIPS2014}
Goodfellow, I., Pouget-Abadie, J., Mirza, M., Xu, B., Warde-Farley, D., Ozair,
  S., Courville, A., Bengio, Y.: Generative adversarial nets. In: {Neural
  Information Processing Systems (NIPS)} (2014)

\bibitem{he2016deep}
He, K., Zhang, X., Ren, S., Sun, J.: Deep residual learning for image
  recognition. In: Proceedings of the IEEE conference on computer vision and
  pattern recognition. pp. 770--778 (2016)

\bibitem{huang2017densely}
Huang, G., Liu, Z., Weinberger, K.Q., van~der Maaten, L.: Densely connected
  convolutional networks. In: Proc CVPR (2017)

\bibitem{krizhevsky2012imagenet}
Krizhevsky, A., Sutskever, I., Hinton, G.E.: Imagenet classification with deep
  convolutional neural networks. In: Advances in neural information processing
  systems. pp. 1097--1105 (2012)

\bibitem{li2017deeper}
Li, D., Yang, Y., Song, Y.Z., Hospedales, T.M.: Deeper, broader and artier
  domain generalization. In: Computer Vision (ICCV), 2017 IEEE International
  Conference on. pp. 5543--5551. IEEE (2017)

\bibitem{MLDG_AAA18}
Li, D., Yang, Y., Song, Y., Hospedales, T.M.: Learning to generalize:
  Meta-learning for domain generalization. In: Conference of the Association
  for the Advancement of Artificial Intelligence (AAAI) (2018)

\bibitem{lin2017feature}
Lin, T.Y., Dollar, P., Girshick, R., He, K., Hariharan, B., Belongie, S.:
  Feature pyramid networks for object detection. In: Proc CVPR (2017)

\bibitem{LongZ0J17}
Long, M., Zhu, H., Wang, J., Jordan, M.I.: Deep transfer learning with joint
  adaptation networks. In: International Conference on Machine Learning (ICML)
  (2017)

\bibitem{mancini2018best}
Mancini, M., Bul{\`o}, S.R., Caputo, B., Ricci, E.: Best sources forward:
  domain generalization through source-specific nets. arXiv preprint
  arXiv:1806.05810  (2018)

\bibitem{MassiRAL}
Mancini, M., Bulo, S.R., Caputo, B., Ricci, E.: Robust place categorization
  with deep domain generalization. IEEE Robotics and Automation Letters  (2018)

\bibitem{Mansour_NIPS09}
Mansour, Y., Mohri, M., Rostamizadeh, A.: Domain adaptation with multiple
  sources. In: {Neural Information Processing Systems (NIPS)} (2009)

\bibitem{ILSVRC15}
Russakovsky, O., Deng, J., Su, H., Krause, J., Satheesh, S., Ma, S., Huang, Z.,
  Karpathy, A., Khosla, A., Bernstein, M., Berg, A.C., Fei-Fei, L.: {ImageNet
  Large Scale Visual Recognition Challenge}. International Journal of Computer
  Vision (IJCV)  \textbf{115}(3),  211--252 (2015).
  \doi{10.1007/s11263-015-0816-y}

\bibitem{russo17sbadagan}
Russo, P., Carlucci, F.M., Tommasi, T., Caputo, B.: From source to target and
  back: symmetric bi-directional adaptive gan. In: Computer Vision and Pattern
  Recognition (CVPR) (2018)

\bibitem{Saenko:2010}
Saenko, K., Kulis, B., Fritz, M., Darrell, T.: Adapting visual category models
  to new domains. In: European Conference on Computer Vision, (ECCV) (2010)

\bibitem{sankaranarayanan2017generate}
Sankaranarayanan, S., Balaji, Y., Castillo, C.D., Chellappa, R.: Generate to
  adapt: Aligning domains using generative adversarial networks. In: Computer
  Vision and Pattern Recognition (CVPR) (2018)

\bibitem{appleGAN}
Shrivastava, A., Pfister, T., Tuzel, O., Susskind, J., Wang, W., Webb, R.:
  Learning from simulated and unsupervised images through adversarial training.
  In: Computer Vision and Pattern Recognition (CVPR) (2017)

\bibitem{Sun:CORAL:AAAI16}
Sun, B., Feng, J., Saenko, K.: Return of frustratingly easy domain adaptation.
  In: Conference of the Association for the Advancement of Artificial
  Intelligence (AAAI) (2016)

\bibitem{Tzeng_ICCV2015}
Tzeng, E., Hoffman, J., Darrell, T., Saenko, K.: Simultaneous deep transfer
  across domains and tasks. In: International Conference in Computer Vision
  (ICCV) (2015)

\bibitem{Hoffman:Adda:CVPR17}
Tzeng, E., Hoffman, J., Darrell, T., Saenko, K.: Adversarial discriminative
  domain adaptation. In: Computer Vision and Pattern Recognition (CVPR) (2017)

\bibitem{venkateswara2017Deep}
Venkateswara, H., Eusebio, J., Chakraborty, S., Panchanathan, S.: Deep hashing
  network for unsupervised domain adaptation. In: ({IEEE}) Conference on
  Computer Vision and Pattern Recognition ({CVPR}) (2017)

\bibitem{cocktail_CVPR18}
Xu, R., Chen, Z., Zuo, W., Yan, J., Lin, L.: Deep cocktail network:
  Multi-source unsupervised domain adaptation with category shift. In: Computer
  Vision and Pattern Recognition (CVPR) (2018)

\bibitem{yu2017deep}
Yu, F., Wang, D., Darrell, T.: Deep layer aggregation. arXiv preprint
  arXiv:1707.06484  (2017)

\bibitem{MDAN_ICLRW18}
Zhao, H., Zhang, S., Wu, G., {a}o P.~Costeira, J., Moura, J.M.F., Gordon, G.J.:
  Multiple source domain adaptation with adversarial learning. In: Workshop of
  the International Conference on Learning Representations (ICLR-W) (2018)

\end{thebibliography}

\end{document}